\documentclass[11pt]{article}
\usepackage{acl2001,times}
\usepackage{epsfig}
\title{Looking Under the Hood: Tools for Diagnosing Your Question
Answering Engine{\footnotesize$^\textrm{1}$}\footnotemark[0]} 

\author{Eric Breck$^{\dagger}$, Marc Light$^{\dagger}$, Gideon
  S. Mann$^{\diamondsuit}$, Ellen Riloff$^{\circ}$, \\ {\bf Brianne 
  Brown$^{\ddagger}$, Pranav Anand$^{*}$, Mats Rooth$^{\mp}$,
  Michael Thelen$^{\circ}$} \\   
 ~ \\
\small
$^{\dagger}$ The MITRE Corporation, 202 Burlington Rd.,Bedford, MA
  01730, \{ebreck,light\}@mitre.org \\
\small
$^{\diamondsuit}$ Department of Computer Science, Johns Hopkins
  University, Baltimore, MD 21218, gsm@cs.jhu.edu \\
\small
$^{\circ}$ School of Computing, University of Utah, Salt Lake City, UT
  84112, \{riloff,thelenm\}@cs.utah.edu \\
\small
$^{\ddagger}$ Bryn Mawr College, Bryn Mawr, PA 19010,
  bbrown@brynmawr.edu\\
\small
$^{*}$ Department of Mathematics, Harvard University, Cambridge, MA
  02138, anand@fas.harvard.edu  \\
\small
$^{\mp}$ Department of Linguistics, Cornell University, Ithaca, NY
  14853, mr249@cornell.edu
\normalsize
}

\date{}

\begin{document}
\maketitle 
\begin{abstract}

\footnotetext[1]{This paper contains a revised Table 2 replacing the one appearing in the Proceedings of the Workshop on Open-Domain
Question Answering, Toulouse, France 2001.}
\setcounter{footnote}{1 }

  In this paper we analyze two question answering tasks : the TREC-8
  question answering task and a set of reading comprehension exams.
  First, we show that Q/A systems perform better when there are
  multiple answer opportunities per question.  Next, we analyze common
  approaches to two subproblems: term overlap for answer sentence
  identification, and answer typing for short answer extraction. We
  present general tools for analyzing the strengths and limitations of
  techniques for these subproblems. Our results quantify the
  limitations of both term overlap and answer typing to distinguish
  between competing answer candidates.

\end{abstract}

\section{Introduction}

When building a system to perform a task, the most important statistic
is the performance on an end-to-end evaluation.  For the task of
open-domain question answering against text collections,
there have been two large-scale end-to-end evaluations:
\cite{trec8-proceedings} and \cite{trec9-proceedings}.  In addition, a
number of researchers have built systems to take reading comprehension
examinations designed to evaluate children's reading
levels \cite{charniak-readcomp,hirschman99,ng2000,riloff-quarc,harper-readcomp}.
The performance statistics have
been useful for determining how well techniques work.

However, raw performance statistics are not enough.  If the score is
low, we need to understand what went wrong
and how to fix it.  If the score is high, it is important to
understand why.  For example, performance may be dependent on
characteristics of the current test set and would not carry over to a
new domain.  It would also be useful to know if there is a particular
characteristic of the system that is central.  If so, then the system
can be streamlined and simplified.

In this paper, we explore ways of gaining insight into question
answering system performance.  First, we analyze the impact of having
multiple answer opportunities for a question. We found that TREC-8 Q/A
systems performed better on questions that had multiple answer
opportunities in the document collection. Second, we present a variety
of graphs to visualize and analyze functions for ranking sentences.
The graphs revealed that relative score instead of absolute score is
paramount. Third, we introduce bounds on functions that use term
overlap\footnote{Throughout the text, we use ``overlap'' to refer to
  the intersection of sets of words, most often the words in the
  question and the words in a sentence.}  to rank sentences.  Fourth,
we compute the expected score of a hypothetical Q/A system that
correctly identifies the answer type for a question and correctly
identifies all entities of that type in answer sentences. We found
that a surprising amount of ambiguity remains because sentences often
contain multiple entities of the same type.

\section{The data}

The experiments in Sections~\ref{ansMult}, \ref{graphs}, and
\ref{bounds} were performed on two question answering data sets: (1)
the TREC-8 Question Answering Track data set and (2) the CBC reading
comprehension data set. We will briefly describe each of these data
sets and their corresponding tasks.

The task of the TREC-8 Question Answering track was to find the answer
to 198 questions using a document collection consisting of roughly
500,000 newswire documents.  For each question, systems were allowed
to return a ranked list of 5 short (either 50-character or
250-character) responses.  As a service to track participants, AT\&T
provided top documents returned by their retrieval engine for each of
the TREC questions.  Sections~\ref{graphs} and \ref{bounds} present
analyses that use all sentences in the top 10 of these documents.
Each sentence is classified as correct or incorrect automatically.
This automatic classification judges a sentence to be correct if it
contains at least half of the stemmed, content-words in the answer
key.  We have compared this automatic evaluation to the TREC-8 QA
track assessors and found it to agree 93-95\% of the time
\cite{breck2000}.

The CBC data set was created for the Johns Hopkins Summer 2000
Workshop on Reading Comprehension.  Texts were collected from the
Canadian Broadcasting Corporation web page for kids
(http://cbc4kids.ca/). They are an average of 24 sentences long.  The
stories were adapted from newswire texts to be appropriate for
adolescent children, and most fall into the following domains:
politics, health, education, science, human interest, disaster,
sports, business, crime, war, entertainment, and environment.  For
each CBC story, 8-12 questions and an answer key were
generated.\footnote{This work was performed by Lisa Ferro and Tim
  Bevins of the MITRE Corporation.  Dr. Ferro has professional
  experience writing questions for reading comprehension exams and led
  the question writing effort.} We used a 650 question subset of the
data and their corresponding 75 stories.  The answer candidates for
each question in this data set were all sentences in the document.
The sentences were scored against the answer key by the automatic
method described previously.

\section{Analyzing the number of answer opportunities per question}
\label{ansMult}

\begin{figure}[htbp]
\centering
\epsfig{figure=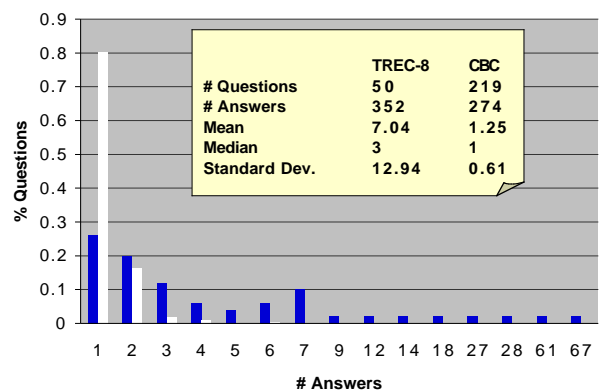,height=2in,width=3.1in}
\caption{Frequency of answers in the TREC-8 (black bars) and CBC 
  (white bars) data sets}
\label{cbc-histograms}
\end{figure}

In this section we explore the impact of multiple answer opportunities
on end-to-end system performance.  A question may have multiple
answers for two reasons: (1) there is more than one different answer
to the question, and (2) there may be multiple instances of each
answer.  For example, {\em ``What does the Peugeot company
manufacture?''} can be answered by {\em trucks}, {\em cars}, or {\em
motors} and each of these answers may occur in many sentences that
provide enough context to answer the question.  The table insert in
Figure~\ref{cbc-histograms} shows that, on average, there are 7 answer
occurrences per question in the TREC-8 collection.\footnote{We would
like to thank John Burger and John Aberdeen for help preparing
Figure~\ref{cbc-histograms}.} In contrast, there are only 1.25 answer
occurrences in a CBC document.  The number of answer occurrences
varies widely, as illustrated by the standard deviations.  The median
shows an answer frequency of 3 for TREC and 1 for CBC, which perhaps
gives a more realistic sense of the degree of answer frequency for
most questions.

To gather this data we manually reviewed 50 randomly chosen TREC-8
questions and identified all answers to these questions in our text
collection. We defined an ``answer'' as a text fragment that contains
the answer string in a context sufficient to answer the question.
Figure~\ref{cbc-histograms} shows the resulting graph.  The $x$-axis
displays the number of answer occurrences found in the text
collection per question and the $y$-axis shows the percentage of
questions that had $x$ answers.  For example, 26\% 
of the TREC-8 questions had
only 1 answer occurrence, and 20\% 
of the TREC-8 questions had exactly 2 answer occurrences (the black
bars).  The most prolific question had 67 answer occurrences (the
Peugeot example mentioned above).  
Figure~\ref{cbc-histograms} also shows the analysis of 219 CBC
questions. In contrast, 80\% 
of the CBC questions had only 1 answer
occurrence in the targeted document, and 16\% 
had exactly 2 answer occurrences.

\begin{figure}[htbp]
\centering
\epsfig{figure=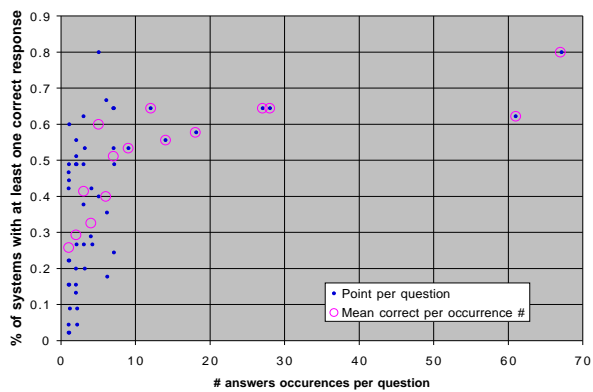,height=2in,width=3.1in}
\caption{Answer repetition vs. system response correctness for TREC-8}
\label{scatter}
\end{figure}

Figure~\ref{scatter} shows the effect that multiple answer
opportunities had on the performance of TREC-8 systems.  Each solid
dot in the scatter plot represents one of the 50 questions we
examined.\footnote{We would like to thank Lynette Hirschman for
suggesting the analysis behind Figure~\ref{scatter} and John Burger
for help with the analysis and presentation.}  The $x$-axis shows the
number of answer opportunities for the question, and the $y$-axis
represents the percentage of systems that generated a correct
answer\footnote{For this analysis, we say that a system generated a
correct answer if a correct answer was in its response set.} for the
question.  E.g., for the question with 67 answer occurrences,
80\% of the systems produced a correct answer.  In
contrast, many questions had a single answer occurrence and the
percentage of systems that got those correct varied from about 2\% to
60\%.

The circles in Figure~\ref{scatter} represent the average percentage
of systems that answered questions correctly for all questions with
the same number of answer occurrences.  For example, on average about
27\% of the systems produced a correct answer for questions that had
exactly one answer occurrence, but about 50\% of the systems produced
a correct answer for questions with 7 answer opportunities.
Overall, a clear pattern emerges: the performance of TREC-8 systems
was strongly correlated with the number of answer opportunities
present in the document collection.

\section{Graphs for analyzing scoring functions of answer candidates}
\label{graphs}

Most question answering systems generate several answer candidates and
rank them by defining a scoring function that maps answer candidates
to a range of numbers.  In this section, we analyze one particular
scoring function: {\em term overlap} between the question and answer
candidate.  The techniques we use can be easily applied to other
scoring functions as well (e.g., weighted term overlap, partial
unification of sentence parses, weighted abduction score, etc.). The
answer candidates we consider are the sentences from the documents.

The expected performance of a system that ranks all sentences using
term overlap is 35\% for the TREC-8 data.  This number is an expected
score because of ties: correct and incorrect candidates may have the
same term overlap score.  If ties are broken optimally, the best
possible score ({\em maximum}) would be 54\%. If ties are broken
maximally suboptimally, the worst possible score ({\em minimum}) would
be 24\%.  The corresponding scores on the CBC data are 58\%
expected, 69\% maximum, and 51\% minimum.  We would like to
understand why the term overlap scoring function works as well as it
does and what can be done to improve it.

\begin{figure}[h]
\centerline{\epsfig{figure=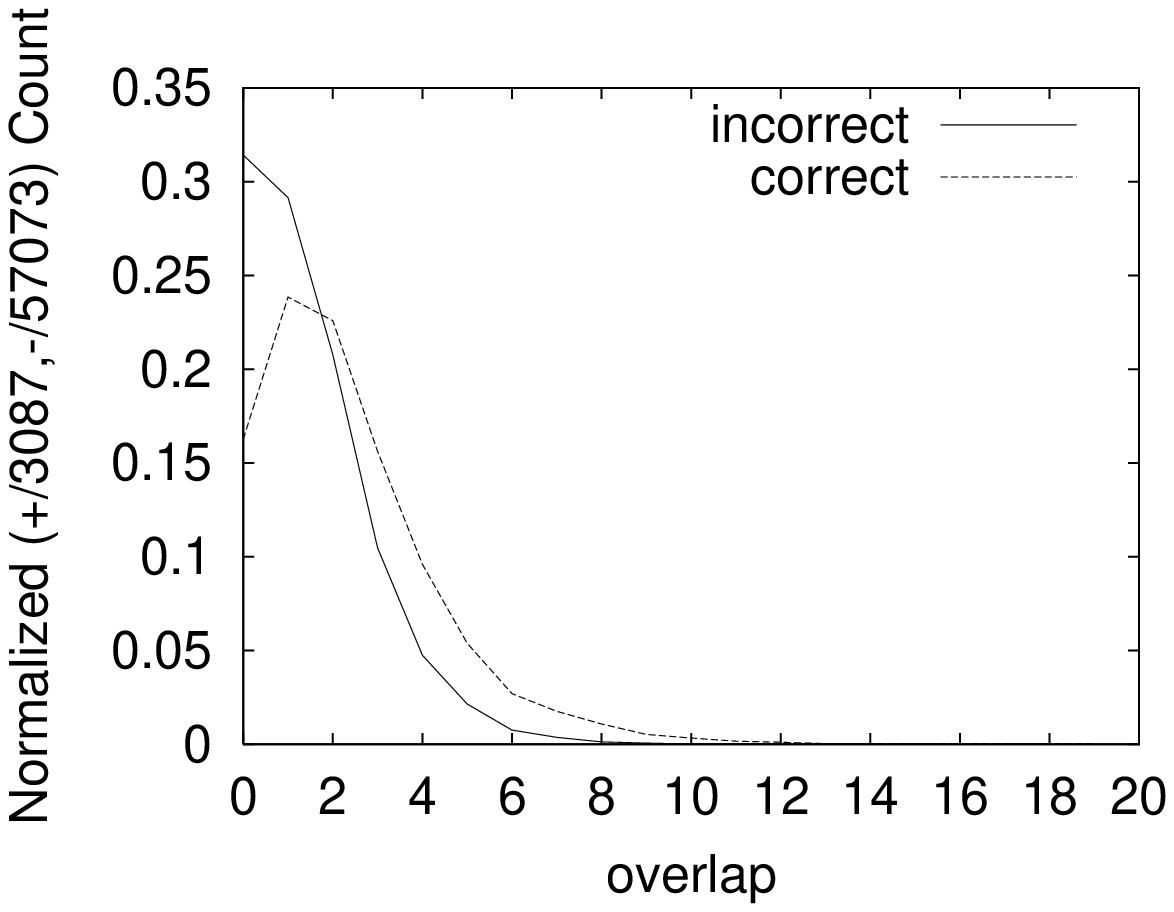,height=2.0in}}
\caption{Pr(overlap=x$\mid$[in]correct) for TREC-8}
\label{camel-overlap-TREC} 
\vspace*{.2in}
\centerline{\epsfig{figure=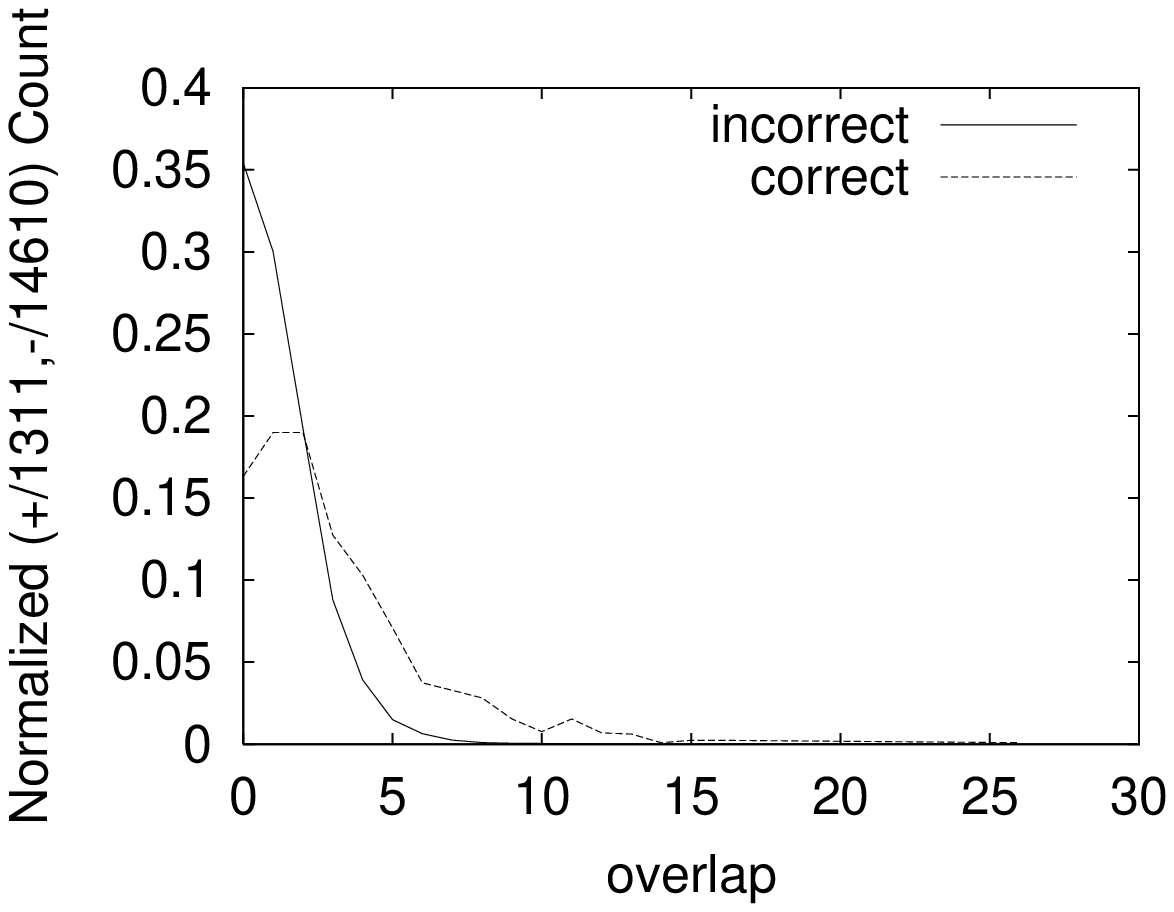,height=2.0in}}
\caption{Pr(overlap=x$\mid$[in]correct) for CBC}
\label{camel-overlap-CBC} 
\end{figure}

Figures~\ref{camel-overlap-TREC} and \ref{camel-overlap-CBC} compare
correct candidates and incorrect candidates with respect to the
scoring function. The $x$-axis plots the range of the scoring
function, i.e., the amount of overlap.  The $y$-axis represents {\bf
  Pr(overlap=x $\mid$ correct)} and {\bf Pr(overlap=x $\mid$
  incorrect)}, where separate curves are plotted for correct and
incorrect candidates.  The probabilities are generated by normalizing
the number of correct/incorrect answer candidates with a particular
overlap score by the total number of correct/incorrect candidates,
respectively.

Figure ~\ref{camel-overlap-TREC} illustrates that the correct
candidates for TREC-8 have term overlap scores distributed between 0 and 10 with
a peak of 24\% at an overlap of 2.  However, the incorrect candidates
have a similar distribution between 0 and 8 with a peak of 32\% at an
overlap of 0.  The similarity of the curves illustrates that it is
unclear how to use the score to decide if a candidate is correct or
not.  Certainly no static threshold above which a candidate is deemed
correct will work.  Yet the expected score of our TREC term overlap system
was 35\%, which is much higher than a random baseline which would get
an expected score of less than 3\% because there are over 40 sentences on
average in newswire documents.\footnote{We also tried dividing the term overlap
  score by the length of the question to normalize for query length
  but did not find that the graph was any more helpful.}

\begin{figure}[h]
\centerline{\epsfig{figure=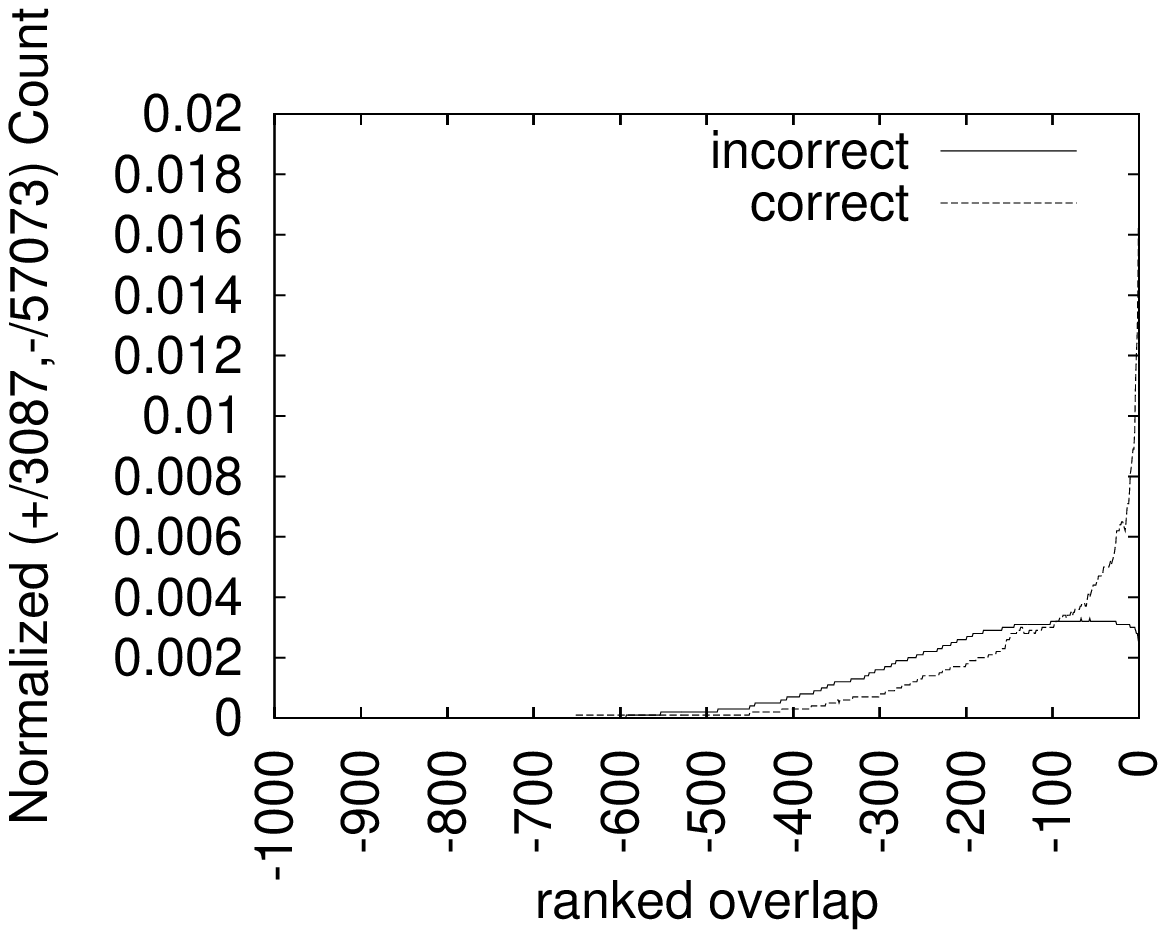,height=2.0in}}
\caption{Pr(rank=x $\mid$ [in]correct) for TREC-8}
\label{camel-overlap-rank-TREC} 
\vspace*{.2in}
\centerline{\epsfig{figure=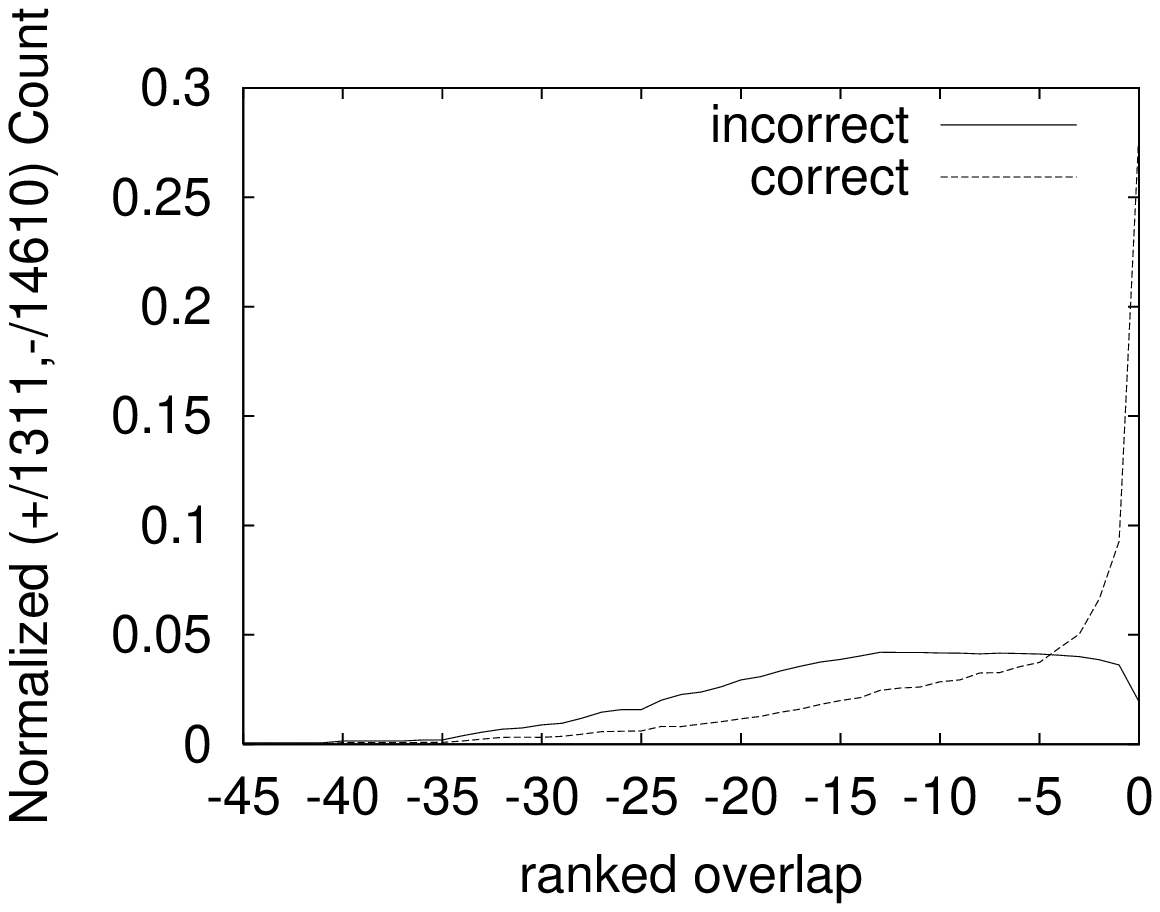,height=2.0in}}
\caption{Pr(rank=x $\mid$ [in]correct) for CBC}
\label{camel-overlap-rank-CBC} 
\end{figure}

After inspecting some of the data directly, we posited that it was not
the absolute term overlap that was important for judging candidate but
how the overlap score compares to the scores of other candidates.  To
visualize this, we generated new graphs by plotting the rank of a
candidate's score on the $x$-axis. For example, the candidate with the
highest score would be ranked first, the candidate with the second
highest score would be ranked second, etc.
Figures~\ref{camel-overlap-rank-TREC} and \ref{camel-overlap-rank-CBC}
show these graphs, which display {\bf Pr(rank=x $\mid$ correct)} and
{\bf Pr(rank=x $\mid$ incorrect)} on the $y$-axis.  The top-ranked
candidate has rank=0.

The ranked graphs are more revealing than the graphs of absolute
scores: the probability of a high rank is greater for correct answers
than incorrect ones.  Now we can begin to understand why the term
overlap scoring function worked as well as it did.  We see that,
unlike classification tasks, there is no good threshold for our
scoring function.  Instead relative score is paramount.  Systems such
as \cite{ng2000} make explicit use of relative rank in their
algorithms and now we understand why this is effective.

\begin{figure}
\centerline{\epsfig{figure=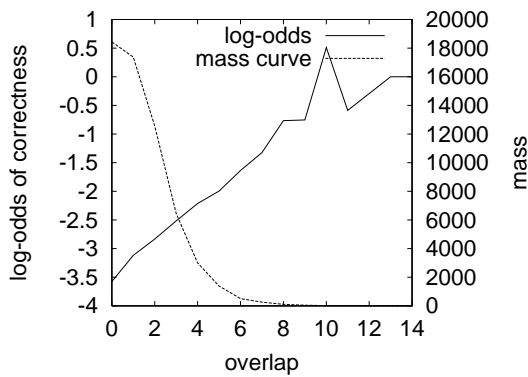,height=2.0in}}
\caption{TREC-8 log odds correct given overlap}
\label{logodds-overlap-TREC} 
\end{figure}

Before we leave the topic of graphing scoring functions, we want to
introduce one other view of the data.
Figure~\ref{logodds-overlap-TREC} plots term overlap scores on the
$x$-axis and the log odds of being correct given a score on the
$y$-axis. The log odds formula is:
\begin{displaymath}
\log\frac{Pr(correct|overlap)}{Pr(incorrect|overlap)}
\end{displaymath}
Intuitively, this graph shows how much more likely a sentence is to be
correct versus incorrect given a particular score.  A second curve,
labeled ``mass,'' plots the number of answer candidates with each
score. Figure~\ref{logodds-overlap-TREC} shows that the odds of being
correct are negative until an overlap of 10, but the mass curve
reveals that few answer candidates have an overlap score greater than
6.

\section{Bounds on scoring functions that use term overlap}
\label{bounds}

The scoring function used in the previous section simply counts the
number of terms shared by a question and a sentence.  One obvious
modification is to weight some terms more heavily than others.  We
tried using inverse document frequence based (IDF) term weighting on
the CBC data but found that it did not improve performance.  The graph
analogous to Figure~\ref{camel-overlap-rank-CBC} but with IDF term
weighting was virtually identical.

Could another weighting scheme perform better?  How well could an
optimal weighting scheme do?  How poorly would the maximally
suboptimal scheme do?  The analysis in this section addresses
these questions.  In essence the answer is the following: the question
and the candidate answers are typically short and thus the number of
overlapping terms is small -- consequently, many candidate answers
have exactly the same overlapping terms and no weighting scheme could
differentiate them.  In addition, subset relations often hold between
overlaps. A candidate whose overlap is a subset of a second
candidate cannot score higher regardless of the weighting
scheme.\footnote{Assuming that all term weights are positive.}  
We formalize these overlap set relations and then calculate statistics
based on them for the CBC and TREC data.

\begin{figure}[htbp]
\fbox{
\begin{minipage}{2.8in}
\footnotesize
Question: How much was Babe Belanger paid to play amateur basketball? \\
\\
S1: She was a member of the winningest \\
\hspace*{.2in} {\bf basketball} team Canada ever had. \\ 
S2: {\bf Babe} {\bf Belanger} never made a cent for her \\
\hspace*{.2in} skills.\\
S3: They were just a group of young women \\ 
\hspace*{.2in} from the same school who liked to \\
\hspace*{.2in}  {\bf play} {\bf amateur} {\bf basketball}. \\ 
S4: {\bf Babe} {\bf Belanger} played with the Grads from  \\
\hspace*{.2in} 1929 to 1937. \\ 
S5: {\bf Babe} never talked about her fabulous career. \\
\hrule
\vspace*{1mm}
MaxOsets : ( \{S2, S4\}, \{S3\} )
\end{minipage}
}
\caption{Example of Overlap Sets from CBC}
\label{qsubset}
\end{figure}

Figure~\ref{qsubset} presents an example from the CBC data.  The four
overlap sets are (i) {\em Babe Belanger}, (ii) {\em basketball}, (iii)
{\em play amateur basketball}, and (iv) {\em Babe}.  In any
term-weighting scheme with positive weights, a sentence containing the
words \textit{Babe Belanger} will have a higher score than sentences
containing just \textit{Babe}, and sentences with \textit{play amateur
  basketball} will have a higher score than those with just
\textit{basketball}.  However, we cannot generalize with respect to
the relative scores of sentences containing \textit{Babe Belanger} and
those containing \textit{play amateur basketball} because some terms
may have higher weights than others.

The most we can say is that the highest scoring candidate must be a
member of $\{S2,S4\}$ or $\{S3\}$. S5 and S1 cannot be ranked highest
because their overlap sets are a proper subset of competing overlap
sets. The correct answer is S2 so an optimal weighting scheme would
have a 50\% chance of ranking S2 first, assuming that it identified
the correct overlap set $\{S2,S4\}$ and then randomly chose between S2
and S4. A maximally suboptimal weighting scheme could rank S2 no lower
than third.

We will formalize these concepts using the following variables:
\begin{quote}
{\em q}: a question (a set of words) \\
{\em s}: a sentence (a set of words) \\
{\em w,v}: sets of intersecting words 
\end{quote}
We define an {\it overlap set} ($o_{w,q}$) to be a set of sentences
(answer candidates) that have the same words overlapping with the
question. We define a {\it maximal overlap set} ($M_q$) as an overlap
set that is not a subset of any other overlap set for the question.
For simplicity, we will refer to a maximal overlap set as a {\it
  MaxOset}.
\begin{itemize}
\item[] $o_{w,q} = \{s| s\cap q = w\}$ 
\item[] $\Omega_{q} = \mbox{all unique overlap sets for } q$ 
\item[] $maximal(o_{w,q})$ ~if~ $\forall o_{v,q} \in \Omega_q, w \not\subset v$ 
\item[] $M_{q} = \{o_{w,q} \in \Omega_{q}\ \mid maximal(o_{w,q})\}$
\item[] $C_{q} = \{s | s \mbox{ correctly answers } q\}$
\end{itemize}

We can use these definitions to give upper and lower bounds on the
performance of term-weighting functions on our two data sets.
Table~\ref{subsetnum} shows the results.  The $max$ statistic is the
percentage of questions for which at least one member of its MaxOsets
is correct.  The $min$ statistic is the percentage of questions for
which all candidates of all of its MaxOsets are correct (i.e., there
is no way to pick a wrong answer).  Finally the $expected max$ is a
slightly more realistic upper bound.  It is equivalent to randomly
choosing among members of the ``best'' maximal overlap set, i.e., the
MaxOset that has the highest percentage of correct members.  Formally,
the statistics for a set of questions $Q$ are computed as:
\begin{displaymath}
 \mbox{max} = \\
 \frac{|\{q| \exists o \in M_q, \exists s \in o \mbox{  s.t.  } s \in
 C_q\}|}{|Q|}  
 \end{displaymath}
\begin{displaymath}
 \mbox{min} = \frac{|\{q|\forall o \in M_q, \forall s \in
 o~~~s \in C_q\}|}{|Q|} 
 \end{displaymath}
\begin{displaymath}
 \mbox{exp. max} = \frac{1}{|Q|}*\sum_{q \in Q} \max_{o \in M_q}
 \frac{|\{s \in o \mbox{ and } s \in C_q\}|}{|o|} 
\end{displaymath}
 
The results for the TREC data are considerably lower than the results
for the CBC data.  One explanation may be that in the CBC data, only
sentences from one document containing the answer are considered.  In
the TREC data, as in the TREC task, it is not known beforehand which
documents contain answers, so irrelevant documents may contain
high-scoring sentences that distract from the correct sentences.

\begin{table}[hbst]
\centering
\begin{tabular}{|l|r|r|r|} \hline
 & exp. max & max & min \\ \hline
CBC training & 72.7\% & 79.0\% & 24.4\% \\
TREC-8 & 48.8\% & 64.7\% & 10.1\% \\ \hline
\end{tabular}
\caption{Maximum overlap analysis of scores}\label{subsetnum}
\end{table}

In Table~\ref{mosbrk}, we present a detailed breakdown of the MaxOset
results for the CBC data.  (Note that the classifications overlap,
e.g., questions that are in ``there is always a chance to get it
right'' are also in the class ``there may be a chance to get it
right.'')  21\% of the questions are literally impossible to
get right using only term weighting because none of the correct
sentences are in the MaxOsets.  
This result illustrates that maximal overlap sets can identify the
limitations of a scoring function by recognizing that some candidates
will \underline{always} be ranked higher than others. Although our
analysis only considered term overlap as a scoring function, maximal
overlap sets could be used to evaluate other scoring functions as
well, for example overlap sets based on semantic classes rather than
lexical items.

\begin{table*}[hbst]
\small
\centerline{\begin{tabular}{|lrr|} \hline
 & \multicolumn{1}{c}{number of} & \multicolumn{1}{c|}{percentage} \\ 
 & \multicolumn{1}{c}{questions}  & \multicolumn{1}{c|}{of questions}\\ \hline
Impossible to get it wrong        &   159 &   24\% \\
  ($\forall o_w \in M_q, \forall s \in o_w, s \in C_q$) & & \\ 
There is always a chance to get it right &   204 &    31\% \\
  ($\forall o_w \in M_q, \exists s \in o_w \mbox{ s.t. } s \in C_q$) &
  & \\ 
There may be a chance to get it right & 514 &   79\% \\
  ($\exists o_w \in M_q \mbox{ s.t. }  \exists s \in o_w \mbox{ s.t. }
  s \in C_q$) & & \\ 
The wrong answers will always be weighted too highly   & 137 &   21\% \\
  ($\forall o_w \in M_q, \forall s \in o_w, s \not\in C_q$) & & \\ 
  There are no correct answers with any overlap with $Q$  &   66 &   10\% \\
  ($\forall s \in d,s $ is incorrect or $s$ has 0 overlap) & & \\ 
  There are no correct answers (auto scoring error)  &   12 &    2\% \\
  ($\forall s \in d,s $ is incorrect) & & \\ \hline
\end{tabular}}
\caption{Maximal Overlap Set Analysis for CBC data}
\label{mosbrk}
\end{table*}

In sum, the upper bound for term weighting schemes is quite low and
the lower bound is quite high.  These results suggest that methods
such as query expansion are essential to increase the feature sets
used to score answer candidates. Richer feature sets could distinguish
candidates that would otherwise be represented by the same features
and therefore would inevitably receive the same score.

\section{Analyzing the effect of multiple answer type occurrences in
  a sentence}
\label{answerType}

In this section, we analyze the problem of extracting short answers
from a sentence.  Many Q/A systems first decide what answer type a
question expects and then identify instances of that type in
sentences.  A scoring function ranks the possible answers using
additional criteria, which may include features of the surrounding
sentence such as term overlap with the question.

For our analysis, we will assume that two short answers that have the
same answer type and come from the same sentence are indistinguishable
to the system. This assumption is made by many Q/A systems: they do
not have features that can prefer one entity over another of the same
type in the same sentence.

We manually annotated data for 165 TREC-9 questions and 186 CBC
questions to indicate perfect question typing, perfect answer
sentence identification, and perfect semantic tagging.
Using these annotations, we measured how much ``answer confusion''
remains if an oracle gives you the correct question type, a sentence
containing the answer, and correctly tags all entities in the sentence
that match the question type.  For example, the oracle tells you that
the question expects a person, gives you a sentence containing the
correct person, and tags all person entities in that sentence. The one
thing the oracle does not tell you is {\it which} person is the
correct one.

Table~\ref{confusability-table} shows the answer types that we used.
Most of the types are fairly standard, except for the {\it Defaultnp}
and {\it Defaultvp} which are default tags for questions
that desire a noun phrase or verb phrase but cannot be more precisely
typed.

We computed an expected score for this hypothetical system as follows:
for each question, we divided the number of correct candidates
(usually one) by the total number of candidates of the same answer
type in the sentence.  For example, if a question expects a {\em
  Location} as an answer and the sentence contains three locations,
then the expected accuracy of the system would be 1/3 because the
system must choose among the locations randomly.  When multiple
sentences contain a correct answer, we aggregated the sentences.
Finally, we averaged this expected accuracy across all questions for
each answer type.

\begin{table}[t]
\footnotesize
\begin{center}
\begin{tabular}{|l|l|c|l|c|} \hline
& \multicolumn{2}{c|}{\bf TREC} & \multicolumn{2}{c|}{\bf CBC} \\
{\it Answer Type} & {\it Score} & {\it Freq} & {\it Score} & {\it Freq} \\ \hline
defaultnp      & .33 & 47 & .25 & 28 \\
organization   & .50 & 1 &  .72 & 3 \\
length         & .50 & 1 &  .75 & 2 \\
thingname      & .58 & 14 & .50 & 1 \\ 
quantity       & .58 & 13 & .77 & 14 \\
agent          & .63 & 19 & .40 & 23 \\
location       & .70 & 24 & .68 & 29  \\
personname     & .72 & 11 & .83 & 13 \\
city           & .73 & 3 &   n/a & 0 \\
defaultvp      & .75 & 2 &  .42 & 15 \\ 
temporal       & .78 & 16 & .75 & 26 \\
personnoun     & .79 & 7 &  .53 & 5 \\
duration       & 1.0 & 3 &  .67 & 4 \\
province       & 1.0 & 2 &  1.0 & 2 \\
area           & 1.0 & 1 &   n/a & 0 \\
day            & 1.0 & 1 &   n/a & 0 \\  
title          & n/a & 0 &  .50 & 1 \\
person         & n/a & 0 &  .67 & 3 \\
money          & n/a & 0 &  .88 & 8 \\
ambigbig       & n/a & 0 &  .88 & 4 \\
age            & n/a & 0 &  1.0 & 2 \\ 
comparison     & n/a & 0 &  1.0 & 1 \\ 
mass           & n/a & 0 &  1.0 & 1 \\ 
measure        & n/a & 0 &  1.0 & 1 \\  \hline
{\bf Overall}    & .59 & 165 &   .61 & 186 \\ \hline
{\bf Overall-dflts} & .69 & 116 &  .70 & 143 \\ \hline
\end{tabular}
\end{center}
\caption{Expected scores and frequencies for each answer type}
\label{confusability-table}
\end{table}

Table~\ref{confusability-table} shows that a system with perfect
question typing, perfect answer sentence identification, and perfect
semantic tagging would still achieve only 59\% accuracy on the TREC-9
data. These results reveal that there are often multiple candidates of
the same type in a sentence.  For example, {\it Temporal} questions
received an expected score of 78\% because there was usually only one
date expression per sentence (the correct one), while {\it Default NP}
questions yielded an expected score of 25\% because there were four
noun phrases per question on average.  Some common types were
particularly problematic.  {\it Agent} questions (most {\em Who}
questions) had an answer confusability of 0.63, while {\it Quantity}
questions had a confusability of 0.58.

The CBC data showed a similar level of answer confusion, with an
expected score of 61\%, although the confusability of individual
answer types varied from TREC. For example, {\it Agent} questions were even
more difficult, receiving a score of 40\%, but {\it Quantity}
questions were easier receiving a score of 77\%.

Perhaps a better question analyzer could assign more specific types to
the {\it Default NP} and {\it Default VP} questions, which skew the
results.  The {\bf Overall-dflts} row of
Table~\ref{confusability-table} shows the expected scores without
these types, which is still about 70\% so a great deal of answer
confusion remains even without those questions.  The confusability
analysis provides insight into the limitations of the answer type set,
and may be useful for comparing the effectiveness of different answer
type sets (somewhat analogous to the use of grammar perplexity in
speech research).

\begin{figure}[htbp]
\fbox{
\begin{minipage}{2.9in}
\footnotesize
Q1: {\it What city is Massachusetts General Hospital located in?}

A1: It was conducted by a cooperative group of oncologists from Hoag,
Massachusetts General Hospital in \underline{{\bf Boston}},
\underline{Dartmouth} College in New Hampshire, UC \underline{San Diego} Medical
Center, McGill University in \underline{Montreal}
and the University of Missouri in \underline{Columbia}. \\

Q2: {\it When was Nostradamus born? }

A2: Mosley said followers of Nostradamus, who lived from
\underline{{\bf 1503}} to \underline{1566},
have claimed ...
\end{minipage}
}
\caption{Sentences with Multiple Items of the Same Type}
\label{multitypes}
\end{figure}

However, Figure~\ref{multitypes} shows the fundamental problem behind
answer confusability. Many sentences contain multiple instances of the
same type, such as lists and ranges. In Q1, recognizing that the
question expects a city rather than a general location is still not
enough because several cities are in the answer sentence. To achieve
better performance, Q/A systems need use features that can more
precisely target an answer.

\section{Conclusion}

In this paper we have presented four analyses of question answering
system performance involving: multiple answer occurence, relative
score for candidate ranking, bounds on term overlap performance, and
limitations of answer typing for short answer extraction.  We hope
that both the results {\em and} the tools we describe will be useful
to others.  In general, we feel that analysis of good performance is
nearly as important as the performance itself and that the analysis of
bad performance can be equally important.

\small
\bibliographystyle{acl}
\bibliography{riloff,hood}
\end{document}